\documentclass[11pt]{article}

\usepackage[final]{acl}
\usepackage{times}
\usepackage{latexsym}
\usepackage[T1]{fontenc}
\usepackage{amsmath,amssymb}
\usepackage{booktabs}
\usepackage{graphicx}
\graphicspath{{figures/}}
\usepackage{float}
\usepackage{microtype}

\hypersetup{colorlinks=true,
  linkcolor=[RGB]{120,20,20},
  citecolor=[RGB]{0,68,120},
  urlcolor=[RGB]{0,68,120},
  pdftitle={Discriminative Axis, Not Data Volume},
  pdfauthor={Abdul Basit Tonmoy}}

\newcommand{\sctask}{\textsc{SpeechCommands}}
\newcommand{\rav}{\textsc{Ravdess}}
\newcommand{\cremad}{\textsc{Crema-D}}

\title{Discriminative Axis, Not Data Volume: What a Contrastive\\
Corpus Teaches an Audio Embedding}

\author{Abdul Basit Tonmoy \\
  Eximius Labs \quad Wabash College \\
  \texttt{atonmoy27@wabash.edu}}

\begin{document}
\maketitle

\begin{abstract}
Scaling the corpus is the default remedy when a contrastive
representation lacks an attribute. We report a case where it does
nothing, and identify what does: adding a lexical-speech round to a
frozen-base multimodal embedding model raises zero-shot keyword spotting
by 76 points while \emph{reducing} speech-emotion recognition by 14.
The loss is not a capacity limit: fine-tuning on 7{,}442 clips from a
prosody-controlled corpus recovers emotion past its pre-speech level at
a five-point keyword cost. Nor is it data volume: 29{,}428 mined clips
whose captions explicitly name emotions, at matched exposure, move
emotion by $-0.0007$. The difference is structural: a contrastive
objective encodes an attribute only when the in-batch negatives cannot
be separated without it; the controlled corpus holds sentence content
fixed, so prosody is the only separating signal, whereas mined captions
name emotion yet remain separable by scene content. Intervention on the
same audio confirms causality: raising caption similarity does not
recover emotion, but collapsing caption diversity so that emotion
becomes the only separating axis recovers it by 8.9 points across three
seeds, with a smaller, same-signed gain on a non-acted corpus, while
keyword accuracy trades back. Corpus structure, not size or caption
vocabulary, controls what a contrastive audio embedding encodes.
\end{abstract}

\begin{figure}[t]
\centering
\includegraphics[width=\columnwidth]{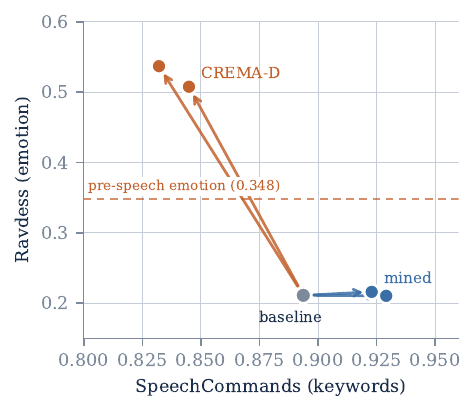}
\caption{Corpus structure, not data volume, decides what a contrastive
audio embedding encodes. Adding a lexical-speech round to a frozen-base
multimodal embedding model raises zero-shot keyword spotting by 76
points ($0.133 \to 0.894$) while reducing emotion recognition by 14
($0.348 \to 0.211$, the grey baseline shown here). Arrows run from that
baseline to each fine-tuned arm: 7{,}442 prosody-controlled clips
(\cremad{}) recover emotion past its pre-speech level (dashed) at a cost
of five keyword points, while a mined corpus four times larger, whose
captions all name emotion explicitly, moves emotion by $-0.0007$ at
matched exposure. Both axes are zero-shot accuracy.}
\label{fig:tradeoff}
\end{figure}

\section{Introduction}

When a learned representation underperforms on some attribute, the
standard response is to add data containing that attribute. We report a
setting where that response fails cleanly: extending the training corpus
of an audio embedding model with large quantities of lexical speech
raised zero-shot keyword spotting by 76 points and \emph{reduced}
speech-emotion recognition by 14 (Figure~\ref{fig:tradeoff}). Both are
speech tasks, evaluated in the same shared space, by the same protocol;
more speech data made the model worse at one of them.

Our setting is an audio embedding built by attaching a trained connector
and modality-gated adapters to a frozen vision-language embedding model~\citep{tonmoy2026fusion}.
We treat the regression as an opportunity rather than a defect, because
it permits a sharper question than ``does more data help'': \emph{which}
property of a contrastive corpus causes an attribute to be encoded? We
eliminate three explanations in turn. The deficit is not capacity:
fine-tuning on 7{,}442 clips from a prosody-controlled corpus restores
emotion past its pre-speech level (Section~\ref{sec:e2}). It is not data
volume: 29{,}428 mined clips whose captions explicitly name emotion, at
matched exposure, move emotion by $-0.0007$ (Section~\ref{sec:e3}). It
is not representational truncation, and the mined fine-tune causes no
collateral damage that could mask an effect (Sections~\ref{sec:e4}
and~\ref{sec:e5}).

What remains is that a contrastive objective is indifferent to
attributes it does not need. An attribute is learned when the in-batch
negatives cannot be separated without it, and ignored when they can,
even if every caption in the corpus names it explicitly. Corpus
construction, not corpus size, decides this; a separability statistic
computable from the corpus alone predicts the outcome
(Section~\ref{sec:sep}), and an intervention on the same audio confirms
the account causally (Section~\ref{sec:e6}).

Our contributions are:
\begin{itemize}\itemsep2pt
\item Large-scale lexical-speech supervision \emph{trades} emotion
  recognition for keyword recognition in a shared audio embedding,
  rather than improving speech understanding uniformly: $+76$ keyword
  points, $-14$ emotion points (Table~\ref{tab:e1},
  Section~\ref{sec:e1}).
\item The loss is not capacity-bound: a corpus $120\times$ smaller than
  the post-speech training corpus restores the lost attribute past its
  original level, $0.211 \to 0.508$ against a pre-speech $0.348$
  (Table~\ref{tab:e2}, Section~\ref{sec:e2}).
\item A matched-exposure negative control shows that four times the
  data, with captions that explicitly name the target attribute, moves
  emotion by $-0.0007$ (Table~\ref{tab:e3}, Section~\ref{sec:e3}).
\item Two further explanations fail: representational truncation
  recovers only 18\% of the gap (Table~\ref{tab:e4},
  Section~\ref{sec:e4}), and a nine-task board shows no general
  degradation that could mask an effect (Table~\ref{tab:board},
  Section~\ref{sec:e5}).
\item A controlled test of the mechanism: restructuring the same corpus
  so that emotion is the required discriminative axis recovers the
  attribute by $+0.089$ across three seeds, whereas raising in-batch
  caption similarity alone does not (Tables~\ref{tab:e6a}
  and~\ref{tab:e6b}, Figure~\ref{fig:e6}, Section~\ref{sec:e6}). This
  converts the account from an observation across corpora into a
  demonstrated cause.
\end{itemize}

\section{Related work}
\label{sec:related}

Our finding sits at the intersection of three lines of work.

\paragraph{Shortcut learning and feature suppression.} A model trained
with a discriminative objective uses whichever feature minimizes the loss
most cheaply, and need not represent alternatives
\citep{geirhos2020shortcut}. In contrastive learning this appears as
\emph{feature suppression}: \citet{chen2021intriguing} construct datasets
with explicitly competing features and show that a few bits of
easy-to-learn shared feature can suppress, and even fully prevent, the
learning of competing ones. \citet{robinson2021shortcut} show the same
tendency follows from optimizing the InfoNCE objective itself, and
propose implicit feature modification to counter it.

Both establish the phenomenon by manipulating the \emph{inputs}
(synthetic competing features, or altered positives and negatives), and
recent work removes it algorithmically: \citet{zhang2024mcl} mitigate
suppression with feature-aware negative sampling that draws an anchor's
negatives from its own cluster, and \citet{bleeker2024shortcuts} reduce
vision-language shortcuts with auxiliary reconstruction objectives. These
treat suppression as a property of the sampler or loss to engineer away on
a fixed dataset. Our contribution is complementary and, to our knowledge,
not made elsewhere: the same suppression is governed by the ordinary
construction of a real training \emph{corpus}, at scale; it is not
remedied by adding data in which the suppressed attribute is explicitly
described; and it is reversed by a structural change to the corpus rather
than to the algorithm. We do not claim the phenomenon of suppression is
new. We isolate the corpus as its operative variable with a
matched-exposure control, and we do so in audio, where CLAP-family corpora
are usually discussed only in terms of scale and caption quality.

\paragraph{Hard negatives and batch composition.} A related line
improves contrastive representations by changing which negatives appear
in a batch, on the argument that easy negatives supply little gradient
signal \citep{robinson2021hard}. That work treats batch composition as an
optimization device for a fixed target. We make a different claim: batch
composition determines \emph{which attribute is learned at all}, not
merely how efficiently a fixed objective converges. The prediction in
Section~\ref{sec:mech} connects the two, since it proposes recovering an
attribute purely by regrouping items already present in the corpus.

\paragraph{Audio--text contrastive embeddings.} Models in the CLAP family
align audio with free-form text and are evaluated zero-shot by embedding
candidate class names \citep{elizalde2023clap,wu2023laionclap}. Corpus
construction in this literature is discussed largely in terms of scale,
caption quality, and noise filtering; \citet{mei2023wavcaps}, for
instance, addresses caption quality at scale. Emotion in particular has
been built into CLAP-style models with explicit attribute supervision
\citep{pan2023gemoclap}; our question is orthogonal, namely when generic
emotion-captioned data does and does not cause emotion to be encoded at
all. Probing studies of what a fixed CLAP embedding already contains
recover low-level acoustic attributes such as loudness and pitch
\citep{martel2026probing}; we instead intervene on the corpus and study a
higher-level attribute whose encoding we show to be corpus-structure
dependent. To our knowledge the structural property we isolate, whether
a corpus makes an attribute \emph{necessary} for discrimination, has
not been separated from scale and caption content with a matched-exposure
control.

We do not claim that attribute competition is novel. The contribution is
the controlled comparison that isolates corpus structure as the operative
variable, in a setting where the intuitive remedy (more data, explicitly
captioned) demonstrably fails.

\section{Setup}
\label{sec:setup}

\paragraph{Model.} A 2B-parameter frozen vision--language embedding
model with a trained audio connector (a query-based resampler) and
modality-gated deep adapters attached to each decoder layer. The gate is
closed on non-audio inputs, so text, image, and video paths execute the
base computation graph unchanged. All training in this paper updates only
the connector and adapters; the base weights are never modified. The
embedding is Matryoshka-nested \citep{kusupati2022matryoshka} with a
default read-out of 1024
dimensions.

\paragraph{Pretraining corpus.} 888{,}800 clips across eleven sources.
The \emph{lexical-speech round} introduced for this study contributes
$\sim$150K spoken-word clips, 70K read-speech clips, and a read-audiobook
set; the remaining sources are environmental audio and audio--caption
pairs. Pretraining uses a frozen-text memory bank, false-negative
masking, and soft labels.

\paragraph{Fine-tuning.} All fine-tunes in Sections~\ref{sec:e2}
and~\ref{sec:e3} use an identical recipe, differing only in the corpus
and the step count: batch 128, no gradient accumulation, peak learning
rate $3{\times}10^{-5}$, false-negative masking and soft labels disabled,
initialized from the same post-speech checkpoint. Holding the recipe
fixed is what makes the two corpora comparable.

\paragraph{Evaluation.} We report two zero-shot classification tasks from
the MAEB audio embedding benchmark in the \textsc{mteb} framework
\citep{muennighoff2022mteb}: \sctask{}
\citep{warden2018speechcommands} (35-way keyword spotting)
and \rav{} (speech emotion). Both are scored by embedding the audio and
the candidate class texts into the shared space and matching; neither is
fine-tuned on. Section~\ref{sec:e5} additionally reports a nine-task
board spanning retrieval, classification, clustering, and reranking, to
test for collateral effects.

\section{Lexical speech trades emotion for keywords}
\label{sec:e1}

Adding the lexical-speech round raises keyword spotting by 76 points and
lowers emotion recognition by 14. Table~\ref{tab:e1} reports the two
tasks before and after the round is added to the pretraining corpus.

\begin{table}[t]
\centering\small
\begin{tabular}{lcc}
\toprule
 & \sctask{} & \rav{} \\
\midrule
pre-speech  & 0.133 & \textbf{0.348} \\
post-speech & \textbf{0.894} & 0.211 \\
\midrule
$\Delta$ & $+0.761$ & $-0.137$ \\
\bottomrule
\end{tabular}
\caption{The lexical-speech round trades one speech attribute for
another: keyword spotting improves by 76 points while emotion
recognition regresses by 14. Both tasks are zero-shot; bold marks the
better value in each column.}
\label{tab:e1}
\end{table}

The keyword gain is unsurprising: the spoken-word corpus pairs isolated
words with their transcriptions, which closely matches the form of the
evaluation. The emotion regression is the finding. Both tasks are speech;
an account in which ``speech data improves speech understanding''
predicts the wrong sign for one of them.

Two explanations are available. Either the model lacks the capacity to
represent both lexical content and prosody and the new supervision
consumed it, or prosody was never required by the objective and the new
supervision actively selected against it. The next two sections separate
these.

\section{The deficit is not capacity}
\label{sec:e2}

A corpus $120\times$ smaller than the post-speech training corpus
restores the lost attribute past its pre-speech level, so the deficit is
not a capacity limit. \cremad{} \citep{cao2014cremad} contains 7{,}442
clips in which 91 actors speak \emph{twelve fixed sentences} across six
emotions. Because lexical content is held approximately constant by
construction, two clips drawn into the same batch typically differ in
speaker and emotion but not in what is said.

\begin{figure*}[t]
\centering
\includegraphics[width=\textwidth]{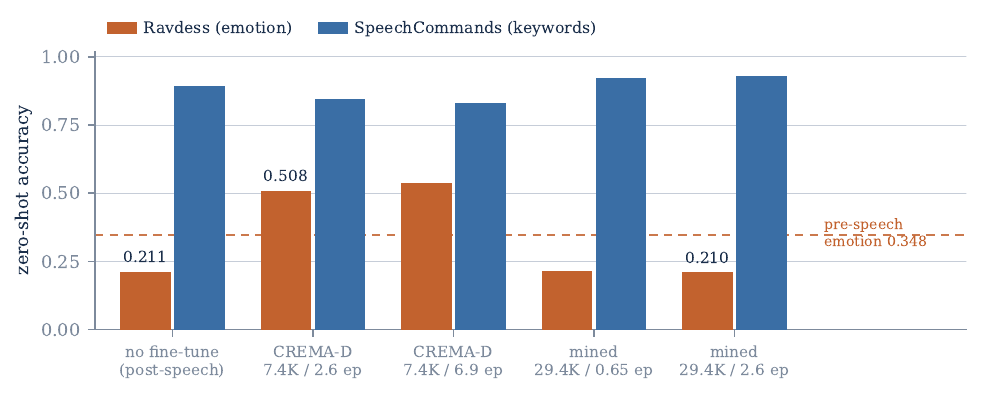}
\caption{Only the prosody-controlled corpus recovers emotion. Both
fine-tuning corpora are shown at two exposures each, against the
post-speech baseline: \cremad{} (7{,}442 clips) lifts emotion past its
pre-speech level of 0.348 (dashed) at a modest keyword cost, while the
mined corpus (29{,}428 clips, every caption naming emotion explicitly)
leaves emotion unchanged and raises keyword accuracy. Both metrics are
zero-shot accuracy and share one axis.}
\label{fig:arms}
\end{figure*}

We fine-tune the post-speech model on \cremad{} for 150 and 400 steps
(Table~\ref{tab:e2}, Figure~\ref{fig:arms}).

\begin{table}[t]
\centering\small
\setlength{\tabcolsep}{3pt}
\begin{tabular}{lrrcc}
\toprule
fine-tune & clips & ep. & \sctask{} & \rav{} \\
\midrule
none        & --      & --  & 0.894 & 0.211 \\
\cremad{}   & 7{,}442 & 2.6 & 0.845 & 0.508 \\
\cremad{}   & 7{,}442 & 6.9 & 0.832 & \textbf{0.537} \\
\bottomrule
\end{tabular}
\caption{A prosody-controlled corpus $120\times$ smaller than the
post-speech training corpus restores emotion past its pre-speech level
of 0.348, at a cost of five points of keyword accuracy. The \cremad{}
rows are 150- and 400-step fine-tunes of the post-speech model.}
\label{tab:e2}
\end{table}

Emotion recovers to 0.508 after 150 steps, exceeding the pre-speech value
of 0.348, while keyword accuracy falls only to 0.845, still more than
six times the pre-speech keyword score. The exchange rate is roughly six
emotion points per keyword point at 150 steps, degrading to about two
between 150 and 400, so the shorter schedule is the better operating
point.

This rules out the capacity account. The model can represent prosody; a
corpus two orders of magnitude smaller than the post-speech training
corpus is sufficient to reinstate it. What the lexical-speech round did
was not exhaust capacity but fail to require prosody.

\section{The deficit is not data volume}
\label{sec:e3}

Four times the data, with captions that explicitly name the target
attribute, moves emotion by $-0.0007$ (Table~\ref{tab:e3}); the deficit
is not data volume either. If prosody merely needed to be \emph{present}
in supervision, then any corpus whose captions describe vocal emotion
should serve, and more of it should serve better. We test this directly.

From the permissive corpora the model already trains on, we mine every
clip whose caption names vocal emotion or paralinguistic vocalization,
using a keyword list covering affect terms and vocalizations, and
excluding obvious non-vocal senses. This yields 29{,}428 clips
(Table~\ref{tab:mined}) with 25{,}392 distinct captions, four times
\cremad{}'s size, drawn from seven sources. Caption diversity is far
higher than \cremad{}'s ($\sim$1.2 clips per caption versus $\sim$20), so
in-batch false negatives are less of a concern here, not more.

\begin{table}[t]
\centering\small
\begin{tabular}{lr}
\toprule
source & clips \\
\midrule
WavCaps / AudioSet-SL & 11{,}040 \\
AudioCaps             & 5{,}691 \\
AudioCaps 2.0         & 5{,}040 \\
FreeSound             & 4{,}317 \\
FSD50K                & 1{,}764 \\
FreeSound (tail)      & 1{,}188 \\
BBC SFX               & 388 \\
\midrule
total                 & 29{,}428 \\
distinct captions     & 25{,}392 \\
\bottomrule
\end{tabular}
\caption{The mined emotion corpus is four times \cremad{}'s size and far
more diverse: 29{,}428 clips from seven sources, 25{,}392 distinct
captions, every caption naming vocal emotion or a paralinguistic
vocalization.}
\label{tab:mined}
\end{table}

\begin{table}[t]
\centering\small
\setlength{\tabcolsep}{3pt}
\begin{tabular}{lrrcc}
\toprule
fine-tune & clips & ep. & \sctask{} & \rav{} \\
\midrule
none   & --       & --   & 0.894 & 0.211 \\
mined  & 29{,}428 & 0.65 & 0.923 & 0.216 \\
mined  & 29{,}428 & 2.6  & 0.929 & 0.210 \\
\bottomrule
\end{tabular}
\caption{Mined emotion-captioned audio leaves emotion unchanged:
$-0.0007$ against the baseline at 2.6 epochs, the exposure at which
\cremad{} produced $+0.297$ (Table~\ref{tab:e2}).}
\label{tab:e3}
\end{table}

The second mined row of Table~\ref{tab:e3} is the control that matters. At 2.6 epochs (exactly
the exposure at which \cremad{} produced $+0.297$) the mined
corpus produces $-0.0007$. Four times the data, captions that explicitly
name the target attribute, an identical recipe, matched optimization, and
no effect.

Keyword accuracy meanwhile \emph{rises} to 0.929. The mined set is
largely human vocalization, so it functions as further vocal-content
supervision. This is consistent with the same competition that produced
the original regression in Section~\ref{sec:e1}: the corpus pushes on the
axis the model already favors (Figure~\ref{fig:tradeoff}).

\section{The deficit is not truncation}
\label{sec:e4}

Evaluating at full dimensionality recovers only 18\% of the emotion gap,
so the deficit is not information hidden in the truncated rungs. Because
the embedding is Matryoshka-nested, a deficit measured at the default
rung could in principle reflect information that is present but
truncated. Table~\ref{tab:e4} evaluates \rav{} at the default and at
full dimensionality for both the pre- and post-speech models.

\begin{table}[t]
\centering\small
\begin{tabular}{lcc}
\toprule
\rav{} & $d{=}1024$ & $d{=}2048$ \\
\midrule
pre-speech  & 0.348 & 0.337 \\
post-speech & 0.211 & 0.224 \\
\midrule
gap & 0.137 & 0.113 \\
\bottomrule
\end{tabular}
\caption{Full dimensionality recovers about 18\% of the gap (0.137 to
0.113) and slightly \emph{lowers} the pre-speech model. The prosody
information is not present-but-truncated.}
\label{tab:e4}
\end{table}

Doubling the read-out closes 18\% of the gap and reduces the pre-speech
model's own score, which is not the signature of information waiting in
the upper rungs.

\section{The intervention is otherwise inert}
\label{sec:e5}

The mined fine-tune leaves a nine-task board essentially unchanged, so
its null effect on emotion is not masked by broad damage. The remaining
alternative was that the fine-tune degrades the representation
generally; Table~\ref{tab:board} rules this out.

\begin{table}[t]
\centering\small
\begin{tabular}{lccc}
\toprule
task & base & mined & $\Delta$ \\
\midrule
\sctask{}            & 0.894 & 0.929 & $+0.035$ \\
MACS T2A             & 0.137 & 0.150 & $+0.013$ \\
GTZAN reranking      & 0.717 & 0.721 & $+0.005$ \\
Clotho T2A           & 0.300 & 0.301 & $+0.002$ \\
VehicleSound clust.  & 0.025 & 0.025 & $+0.001$ \\
\rav{}               & 0.211 & 0.210 & $-0.001$ \\
UrbanSound8K T2A     & 0.010 & 0.009 & $-0.001$ \\
BeijingOpera         & 0.924 & 0.919 & $-0.004$ \\
GTZAN genre          & 0.644 & 0.635 & $-0.009$ \\
\bottomrule
\end{tabular}
\caption{The mined fine-tune is not broadly destructive. Across a
nine-task board, no cell is degraded by more than $0.009$; the only
gains above that are keyword spotting ($+0.035$) and MACS text-to-audio
retrieval ($+0.013$). Rows are ordered by $\Delta$.}
\label{tab:board}
\end{table}

The mined corpus is therefore not harming the representation. It simply
does not teach prosody.

\begin{figure*}[t]
\centering
\includegraphics[width=\textwidth]{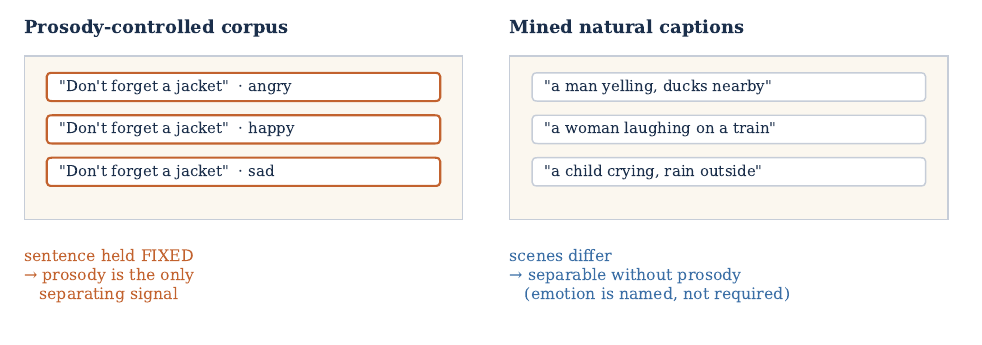}
\caption{An attribute is encoded only when the negatives cannot be
separated without it. Left: in the prosody-controlled corpus, lexical
content is held fixed, so prosody is the only remaining signal and the
objective cannot be minimized without encoding it. Right: in mined
natural captions, scenes differ, so items are separable by scene content
alone; emotion is \emph{named} in the caption but never \emph{required}.}
\label{fig:mech}
\end{figure*}

\section{Caption separability predicts what is learned}
\label{sec:sep}

A text-side statistic, computed from the corpus alone with no training,
orders the corpora exactly as their outcomes do. The account above
implies such a property: if the text side alone resolves an anchor from
its negatives, the audio side is never forced to carry the attribute. We
therefore measure, directly from the cached text embeddings, how
confusable an item's caption is with the captions of its negatives.

For sampled negative sets of 128 items we report two statistics
(Table~\ref{tab:sep}): \textbf{collision rate}, the fraction of anchors
whose \emph{exact} caption also appears on another item in the set; and
\textbf{mean max similarity}, the average cosine similarity between an
anchor's caption and its most similar distractor. Both are properties of
the corpus, not of any trained model.

\begin{table}[t]
\centering\small
\setlength{\tabcolsep}{4pt}
\begin{tabular}{lrrr}
\toprule
corpus & clips/cap. & coll. & max sim \\
\midrule
\cremad{}       & 19.90 & \textbf{0.556} & \textbf{0.971} \\
mined emotion   &  1.16 & 0.038 & 0.827 \\
AudioCaps       &  1.09 & 0.003 & 0.804 \\
spoken words    &  5.53 & 0.006 & 0.759 \\
\bottomrule
\end{tabular}
\caption{Text-side separability singles out the one corpus that taught
prosody. In \cremad{}, 55.6\% of anchors have an exact caption duplicate
among their negatives and the nearest distractor sits at cosine 0.971:
the text side cannot resolve the positive. Every other corpus resolves
easily.}
\label{tab:sep}
\end{table}

\cremad{}, the only corpus that taught prosody, is an outlier by more
than an order of magnitude in collision rate. The mined corpus, whose
captions all name emotion, is 14$\times$ less confusable and did not
teach it. The spoken-word corpus that produced the original regression is
the \emph{least} confusable of all: single-word captions are semantically
far apart, so the text side resolves everything and the audio side need
only encode which word was said.

This inverts a standard quality heuristic. Caption diversity is normally
a virtue, and by that measure \cremad{} is the worst corpus in the table.
It is the only one that works, because low diversity is what forces the
audio side to do the discriminating. We suggest reporting these
statistics alongside corpus size when a contrastive corpus is introduced,
since they are free to compute and predict which attributes the corpus
can teach.

\section{Mechanism}
\label{sec:mech}

We propose that a contrastive objective encodes an attribute only when
its negative set cannot be separated without it (Figure~\ref{fig:mech}).

\paragraph{The negative set is not the batch.} Our objective draws
negatives from the batch \emph{together with} a FIFO memory bank of
frozen text embeddings, so the relevant comparison set is far larger than
a single batch and is drawn from the corpus at large. This matters for
the argument, and it sharpens rather than weakens it: the question is not
whether 128 sampled items happen to be confusable, but whether the
\emph{corpus} supplies text-side negatives that make the attribute
necessary. A corpus with few distinct captions is confusable everywhere,
not just occasionally.

In \cremad{}, sentence content is fixed by construction, and the corpus
contains only 374 distinct captions across 7{,}442 clips. Any negative
drawn from the bank is therefore likely to carry near-identical text to
the anchor. The text side cannot resolve the positive, so the loss is not
minimizable without representing prosody. The attribute is
\emph{required}.

In the mined corpus, an item such as ``a man yelling and ducks in the
background'' does name the emotional attribute, but its 25{,}392 distinct
captions mean almost every negative differs in scene content. The item is
separable by scene alone (ducks, vehicles, water), so the loss
reaches its minimum without ever encoding how the voice sounds. The
attribute is \emph{mentioned} but not \emph{required}.

Note that this makes caption diversity, ordinarily a marker of corpus
quality, the very property that lets the model skip the attribute.
Section~\ref{sec:sep} makes that quantitative.

This account explains the failure of scale in Section~\ref{sec:e3}, the
success of a much smaller corpus in Section~\ref{sec:e2}, and the
original regression in Section~\ref{sec:e1}: a corpus of isolated spoken
words makes lexical identity the discriminative axis, so prosody is not
merely unsupervised but selected against.

\paragraph{A testable prediction.} If the mechanism is correct, the mined
corpus should encode emotion once emotion is \emph{made} the residual
discriminative axis, with no new audio. Section~\ref{sec:e6} tests this
directly, first under the most literal reading (regroup the batches) and
then under the sharper one the mechanism actually implies (remove the
text-side shortcut so the negatives cannot be separated without emotion).

\section{Testing the mechanism}
\label{sec:e6}

\begin{figure}[t]
\centering
\includegraphics[width=\columnwidth]{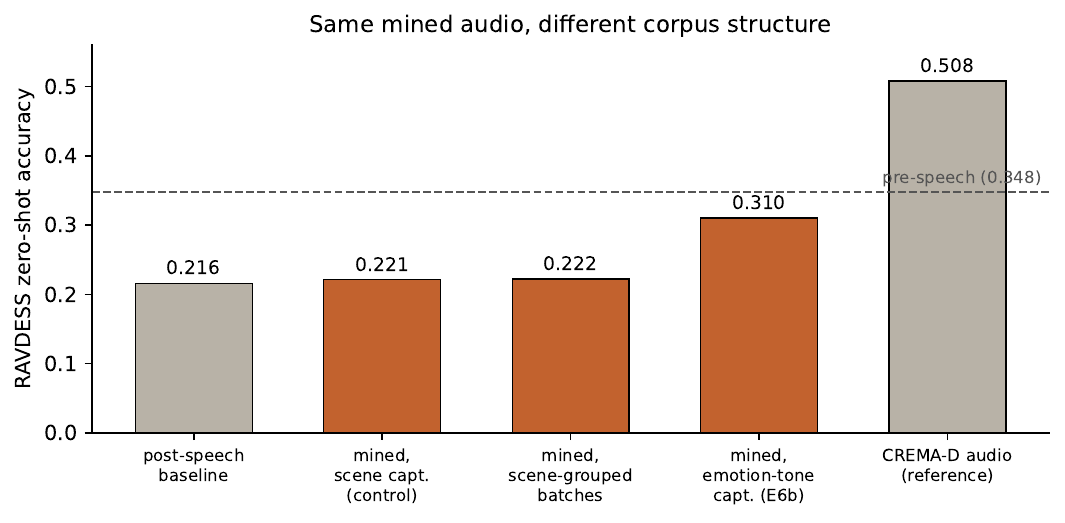}
\caption{Caption collision, not caption similarity, recovers emotion.
The same mined audio is trained under four corpus structures (\rav{}
zero-shot emotion): grouping batches by scene similarity (E6a) does not
move emotion, while collapsing caption diversity so that emotion is the
required discriminative axis (E6b) recovers it most of the way to the
pre-speech level (dashed), with none of the audio changed. \cremad{}, a
different, prosody-controlled audio corpus, is shown for reference.}
\label{fig:e6}
\end{figure}

Restructuring the negatives of the mined corpus recovers emotion with no
new audio, exactly as the mechanism predicts: emotion returns when
prosody becomes the only way to tell training items apart, and not
before. We test this on the
\emph{identical} mined clips, holding the fine-tuning recipe of
Section~\ref{sec:setup} fixed and changing only the structure of the
contrastive negatives. Both interventions initialize from the post-speech
model and run to the same 2.6-epoch exposure as Section~\ref{sec:e3}.
Figure~\ref{fig:e6} summarizes the outcome.

\subsection{Similarity is not enough}
\label{sec:e6a}

Scene-grouped batching raises caption similarity and leaves emotion
unmoved. The most literal reading of the mechanism is to group each batch by scene. We order the
corpus so that every batch of 128 is a nearest-neighbor cluster in the
space of emotion-stripped caption embeddings, which raises the mean
within-batch caption similarity from $0.65$ to $0.71$ (nearest-distractor
cosine $0.83 \to 0.89$) while keeping $6.6$ emotions per batch on average,
and we draw negatives from the batch alone. Table~\ref{tab:e6a} shows the
result: scene-grouped batching does not recover emotion (\rav{} $0.222$,
against a random-batch control of $0.231$ and a pre-intervention baseline
of $0.216$).

\begin{table}[t]
\centering\small
\setlength{\tabcolsep}{4pt}
\begin{tabular}{lcc}
\toprule
in-batch negatives & \rav{} & \sctask{} \\
\midrule
post-speech base & 0.216 & 0.915 \\
random control   & 0.231 & 0.926 \\
scene-grouped    & 0.222 & 0.931 \\
\bottomrule
\end{tabular}
\caption{Scene-grouped batching (E6a) leaves emotion unmoved despite
raising the nearest-distractor caption cosine to $0.89$ (within-batch
mean similarity $0.65 \to 0.71$). Grouping raised similarity but not
\emph{collision}: captions stayed distinct (exact-duplicate rate $0.04$,
versus \cremad{}'s $0.56$), so the text side could still resolve items by
fine caption differences and prosody was never required.}
\label{tab:e6a}
\end{table}

This sharpens the mechanism. The operative property is not proximity but
\emph{collision}: \cremad{} works because many items carry the same
caption, not because their captions are merely similar. Grouping moved the
similarity statistic of Section~\ref{sec:sep} without moving the collision
statistic, and emotion did not move with it.

\subsection{Collision recovers emotion}
\label{sec:e6b}

Collapsing caption diversity so that emotion is the only separating axis
recovers it on the same audio (Table~\ref{tab:e6b}). To remove the
text-side shortcut directly, we relabel every mined clip with a \cremad{}-style emotion-tone caption (for example
``Speech with an angry tone, spoken by a person''), assigning the label
from the clip's dominant emotion keyword, and cache these as the text
targets; the audio is untouched. This gives the corpus \cremad{}'s
structure: caption diversity collapses from $25{,}392$ distinct captions
to $16$, the within-batch collision rate rises from $0.04$ to $0.98$, and
the full-corpus frozen-text bank now makes emotion the only axis that
separates an anchor from its many same-caption negatives. The control is
the identical run on the original scene-descriptive captions.

\begin{table}[t]
\centering\small
\setlength{\tabcolsep}{4pt}
\begin{tabular}{lcc}
\toprule
mined captions & \rav{} & \sctask{} \\
\midrule
scene-descriptive & 0.221 & 0.930 \\
emotion-tone      & \textbf{0.310} & 0.903 \\
\midrule
\cremad{} audio (ref) & 0.508 & 0.845 \\
\bottomrule
\end{tabular}
\caption{Caption collision (E6b): coarsening the mined captions from
$25{,}392$ distinct to $16$ \cremad{}-style emotion-tone labels makes
emotion the required axis on the \emph{same} mined audio, recovering
\rav{} by $+0.089$, most of the way back to the pre-speech level
($0.348$), while keyword accuracy trades back. Three seeds each, random
batching with the full-corpus text bank: control $0.221$
($0.217/0.224/0.222$), collision $0.310$ ($0.308/0.308/0.315$), positive
at every seed with non-overlapping ranges. \cremad{} audio ($374$
captions) is shown for reference; only the caption structure differs
between the two mined rows.}
\label{tab:e6b}
\end{table}

The same clips that produced no effect as scene descriptions
(Section~\ref{sec:e3}) recover emotion once relabeled so that emotion is
required (Table~\ref{tab:e6b}). The effect is large ($+0.089$ \rav{}) and holds at every seed
with non-overlapping ranges, and it is not free: keyword accuracy falls
from $0.930$ to $0.903$, the same competition that produced the original
regression in Section~\ref{sec:e1}, now running in reverse. The recovery
is partial ($0.310$ against \cremad{}'s $0.508$ on the same eval),
consistent with the mined vocalization clips carrying a noisier prosodic
signal than \cremad{}'s controlled recordings.

\paragraph{A second, non-acted benchmark.} On \textsc{Iemocap} \citep{busso2008iemocap} emotion
classification (a different corpus, improvised rather than lexically
matched), the same recovery appears in sign but attenuated: the collision
model scores $0.223$ against the scene-caption control's $0.204$ and the
baseline's $0.201$ (three-seed means; the gap is positive at every seed
with non-overlapping ranges, but roughly a quarter of the \rav{} effect).
The main effect is therefore strong within the acted-emotion paradigm and
directionally, seed-robustly confirmed outside it, not uniform across
emotion corpora.

\section{Discussion}

The practical reading is that ``add data containing the attribute'' is
not a well-formed instruction for contrastive training. The mined corpus
satisfies that instruction completely (every clip's caption names
vocal emotion), and it does nothing. The controlled corpus violates the
usual desiderata for corpus quality, in that it has only 374 distinct
captions across 7{,}442 clips and therefore a high false-negative rate,
and it works.

This inverts the usual quality heuristics. Caption diversity, corpus
size, and label coverage are all proxies for the thing that matters, and
in this case they are anti-correlated with it. What matters is whether
the corpus makes the target attribute necessary for discrimination.

A corollary worth stating: this predicts that mixing a controlled corpus
into a large natural one may dilute exactly the property that makes it
work, since the controlled items would be batched against natural
negatives and become separable without prosody. If so, the intervention
belongs at the batch level rather than the corpus level. We have not
tested this.

\section{Limitations}

The fine-tuning arms of Sections~\ref{sec:e1} through~\ref{sec:e5} are
single runs; only the interventions of Section~\ref{sec:e6} are
replicated across three seeds. The single-run effect sizes are large
relative to typical seed variance in this setup, but we do not report
seed replicates for them, and the quantitative exchange rates in
Section~\ref{sec:e2} should be read as approximate rather than measured.

The emotion axis rests principally on a single benchmark, and the two
corpora share
more structure than is ideal for a clean test. \cremad{} and \rav{} are
both \emph{acted} corpora with categorical labels, and both hold lexical
content approximately fixed across emotions: \citet{livingstone2018ravdess}
use lexically-matched statements by design, exactly the property we argue
makes \cremad{} effective. Section~\ref{sec:e2} is therefore partly
domain transfer, and the recovery may be assisted by the evaluation
sharing the training corpus's structural bias. We therefore report a
second, non-acted emotion benchmark, \textsc{Iemocap}, on which the
structural recovery of Section~\ref{sec:e6} replicates in sign but at a
much smaller magnitude.
The main effect should be read as strong within the acted-emotion
paradigm and directionally confirmed outside it, not as uniform across
emotion corpora; the mined vocalization corpus transfers imperfectly to
conversational emotion.

The keyword gains in Section~\ref{sec:e1} follow from training on
spoken-word data. This is a genuine capability gain but it is not
zero-shot generalization in the strict sense for this model, and we do
not claim it as such.

The mechanism is no longer only an inference from the contrast between
two corpora. Section~\ref{sec:e6} tests it directly on the same audio:
raising in-batch caption similarity leaves emotion unmoved, while
collapsing caption diversity so that emotion is the required
discriminative axis recovers it. We are deliberate about what this does
and does not show. The emotion-tone relabeling is itself a supervised
signal, and a reader could call it a relabeling into a classification
task. The point is that the \emph{same} signal, present as free text in
the mined captions of Section~\ref{sec:e3}, produced no effect, and
became usable only once the corpus structure made it necessary. What the
intervention isolates is that structure, not the mere presence of the
label, is the operative variable. It remains a single base model and a
single audio pipeline; whether the effect is objective-level rather than
architecture-specific rests on the model-agnostic prior results we cite
in Section~\ref{sec:related}.

\paragraph{Data contamination.} Our pretraining corpus includes 70K
Common Voice clips sampled from the \texttt{validated} pool. In Common
Voice, \texttt{validated} is the superset from which train, dev, and test
splits are drawn, and the corpus is cumulative across releases. Replaying
our (deterministic) ingest to recover the exact sampled clip identifiers,
we measure \textbf{44 of 200 (22\%) of the CommonVoice v21 mini test
clips present in our training set}, alongside 47\% verbatim sentence
overlap. The cause is structural rather than accidental: the ingest
streams the archive in order and stops at a target count, archive order
tracks clip identifier, and low identifiers are the older recordings that
the v21 test split draws from, so ``the first $N$ that pass the filters''
systematically over-samples the evaluation era. The effect is compounded
because our spoken-word corpus is itself derived from Common Voice. We
therefore do not report any Common Voice retrieval result for these
models, and the evaluations in this paper are drawn from tasks whose
source corpora we do not train on.

We flag the general hazard, since it is not specific to us: several
corpora in wide use as \emph{training} data (Common Voice, \cremad{},
IEMOCAP, VoxPopuli, FLEURS, GigaSpeech) are simultaneously \emph{tasks}
in audio embedding benchmarks. A model trained on the obvious speech data
is contaminated on several cells by default, and sampling from a pooled
``validated'' split rather than a designated train split makes this
likely rather than possible. Corpora intended for training should record
source clip identifiers so that overlap can be audited after the fact;
ours did not, and recovering them required replaying the ingest.

Finally, \cremad{} is distributed under a share-alike database license
and is used here for controlled experimentation only. No model weights
trained on it are released, and no model trained on it is submitted to a
benchmark whose tasks derive from it.

\section{Conclusion}

Adding a large lexical-speech round to a contrastive audio embedding
improved keyword spotting by 76 points and cost 14 points of emotion
recognition. The lost attribute was recovered by a corpus $120\times$
smaller than the training corpus, whose construction made prosody the
only way to tell training items apart, and was not recovered at all by a
corpus four times larger
whose captions named emotion explicitly but whose items remained
separable without it. A direct intervention closes the loop: restructuring
that same mined corpus so that emotion is the only axis separating
training items recovers it, while raising in-batch similarity alone does
not, so the structural property is a cause and not a correlate. What a
contrastive corpus teaches is decided by what its negatives make
necessary. Data volume and caption vocabulary are poor proxies for that,
and optimizing them is not a substitute for designing it. To give a
contrastive embedding an attribute, construct the corpus so that its
negatives cannot be separated without that attribute.

\bibliography{refs}

\end{document}